\documentclass[11pt,a4paper]{article}
\usepackage[hyperref]{acl2018}
\usepackage{times}
\usepackage{latexsym}

\usepackage{url}

\usepackage{graphicx}
\usepackage{amsmath}
\usepackage{amsfonts}
\usepackage{multirow}

\aclfinalcopy 
\def\aclpaperid{871} 

\title{Aspect Based Sentiment Analysis with Gated Convolutional Networks}

\author{Wei Xue \textnormal{and} Tao Li \\
  School of Computing and Information Sciences \\
  Florida International University, Miami, FL, USA \\
  {\tt {\{wxue004, taoli\}@cs.fiu.edu}}
  }

\date{}

\begin{document}
\maketitle

\begin{abstract}
Aspect based sentiment analysis (ABSA) can provide more detailed information than general sentiment analysis, because it aims to predict the sentiment polarities of the given aspects or entities in text. We summarize previous approaches into two subtasks: aspect-category sentiment analysis (ACSA) and aspect-term sentiment analysis (ATSA). Most previous approaches employ long short-term memory and attention mechanisms to predict the sentiment polarity of the concerned targets, which are often complicated and need more training time.
We propose a model based on convolutional neural networks and gating mechanisms, which is more accurate and efficient. First, the novel Gated Tanh-ReLU Units can selectively output the sentiment features according to the given aspect or entity. The architecture is much simpler than attention layer used in the existing models. Second, the computations of our model could be easily parallelized during training, because convolutional layers do not have time dependency as in LSTM layers, and gating units also work independently.
The experiments on SemEval datasets demonstrate the efficiency and effectiveness of our models.~\footnote{The code and data is available at \url{https://github.com/wxue004cs/GCAE}}
\end{abstract}

\section{Introduction}
Opinion mining and sentiment analysis~\cite{Pang:2008wj} on user-generated reviews can provide valuable information for providers and consumers. Instead of predicting the overall sentiment polarity, fine-grained aspect based sentiment analysis (ABSA)~\cite{Liu:2012ke} is proposed to better understand reviews than traditional sentiment analysis. Specifically, we are interested in the sentiment polarity of aspect categories or target entities in the text. Sometimes, it is coupled with aspect term extractions~\cite{xue2017mtna}. 
A number of models have been developed for ABSA, but there are two different subtasks, namely aspect-category sentiment analysis (ACSA) and aspect-term sentiment analysis (ATSA). The goal of ACSA is to predict the sentiment polarity with regard to the given aspect, which is one of a few predefined categories. On the other hand, the goal of ATSA is to identify the sentiment polarity concerning the target entities that appear in the text instead, which could be a multi-word phrase or a single word. The number of distinct words contributing to aspect terms could be more than a thousand. For example, in the sentence ``\textit{Average to good Thai food, but terrible delivery.}'',  ATSA would ask the sentiment polarity towards the entity \textit{Thai food}; while ACSA would ask the sentiment polarity toward the aspect \textit{service}, even though the word \textit{service} does not appear in the sentence.

Many existing models use LSTM layers~\cite{Hochreiter:1997fq} to distill sentiment information from embedding vectors, and apply attention mechanisms~\cite{Bahdanau:2014vz} to enforce models to focus on the text spans related to the given aspect/entity. Such models include Attention-based LSTM with Aspect Embedding (ATAE-LSTM)~\cite{Wang:2016tf} for ACSA; Target-Dependent Sentiment Classification (TD-LSTM)~\cite{Tang:2016th}, Gated Neural Networks~\cite{Zhang:2016th} and Recurrent Attention Memory Network (RAM)~\cite{Chen:2017wv} for ATSA.
Attention mechanisms has been successfully used in many NLP tasks. It first computes the alignment scores between context vectors and target vector; then carry out a weighted sum with the scores and the context vectors. 
However, the context vectors have to encode both the aspect and sentiment information, and the alignment scores are applied across all feature dimensions regardless of the differences between these two types of information.  
Both LSTM and attention layer are very time-consuming during training. LSTM processes one token in a step. Attention layer involves exponential operation and normalization of all alignment scores of all the words in the sentence~\cite{Wang:2016tf}. 
Moreover, some models needs the positional information between words and targets to produce weighted LSTM~\cite{Chen:2017wv}, which can be unreliable in noisy review text. 
Certainly, it is possible to achieve higher accuracy by building more and more complicated LSTM cells and sophisticated attention mechanisms; but one has to hold more parameters in memory, get more hyper-parameters to tune and spend more time in training.
In this paper, we propose a fast and effective neural network for ACSA and ATSA based on convolutions and gating mechanisms, which has much less training time than LSTM based networks, but with better accuracy.

For ACSA task, our model has two separate convolutional layers on the top of the embedding layer, whose outputs are combined by novel gating units. Convolutional layers with multiple filters can efficiently extract n-gram features at many granularities on each receptive field.
The proposed gating units have two nonlinear gates, each of which is connected to one convolutional layer. With the given aspect information, they can selectively extract aspect-specific sentiment information for sentiment prediction. 
For example, in the sentence ``\textit{Average to good Thai food, but terrible delivery.}'', when the aspect \textit{food} is provided, the gating units automatically ignore the negative sentiment of aspect \textit{delivery} from the second clause, and only output the positive sentiment from the first clause.
Since each component of the proposed model could be easily parallelized, it has much less training time than the models based on LSTM and attention mechanisms. 
For ATSA task, where the aspect terms consist of multiple words, we extend our model to include another convolutional layer for the target expressions. 
We evaluate our models on the SemEval datasets, which contains restaurants and laptops reviews with labels on aspect level. To the best of our knowledge, no CNN-based model has been proposed for aspect based sentiment analysis so far.

\section{Related Work}
We present the relevant studies into following two categories.

\subsection{Neural Networks}
Recently, neural networks have gained much popularity on sentiment analysis or sentence classification task. Tree-based recursive neural networks such as Recursive Neural Tensor Network~\cite{Socher:2013ug} and Tree-LSTM~\cite{Tai:2015wp}, 
make use of syntactic interpretation of the sentence structure, but these methods suffer from time inefficiency and parsing errors on review text. 
Recurrent Neural Networks (RNNs) such as LSTM~\cite{Hochreiter:1997fq} and GRU~\cite{Chung:2014wf}
have been used for sentiment analysis on data instances having variable length~\cite{Tang:2015ts,Xu:2016vb,SiweiLai:2014to}.
There are also many models that use convolutional neural networks (CNNs)~\cite{Collobert:2011tk,Kalchbrenner:2014wl,Kim:2014vt,Conneau:2016to} in NLP, which also prove that convolution operations can capture compositional structure of texts with rich semantic information without laborious feature engineering.

\subsection{Aspect based Sentiment Analysis}
There is abundant research work on aspect based sentiment analysis. Actually, the name ABSA is used to describe two different subtasks in the literature. We classify the existing work into two main categories based on the descriptions of sentiment analysis tasks in SemEval 2014 Task 4~\cite{Pontiki:2014ex}: Aspect-Term Sentiment Analysis and Aspect-Category Sentiment Analysis.

\textbf{Aspect-Term Sentiment Analysis}. In the first category, sentiment analysis is performed toward the aspect terms that are labeled in the given sentence. A large body of literature tries to utilize the relation or position between the target words and the surrounding context words either by using the tree structure of dependency or by simply counting the number of words between them as a relevance information~\cite{Chen:2017wv}.

Recursive neural networks~\cite{Lakkaraju:2014vy,Dong:2014vd,Wang:2016vm} rely on external syntactic parsers, which can be very inaccurate and slow on noisy texts like tweets and reviews, which may result in inferior performance. 

Recurrent neural networks are commonly used in many NLP tasks as well as in ABSA problem. TD-LSTM~\cite{Tang:2016th} and gated neural networks~\cite{Zhang:2016th} use two or three LSTM networks to model the left and right contexts of the given target individually. A fully-connected layer with gating units predicts the sentiment polarity with the outputs of LSTM layers. 

Memory network~\cite{Weston:2014va} coupled with multiple-hop attention attempts to explicitly focus only on the most informative context area to infer the sentiment polarity towards the target word~\cite{Tang:2016uz,Chen:2017wv}. Nonetheless, memory network simply bases its knowledge bank on the embedding vectors of individual words~\cite{Tang:2016uz}, which makes itself hard to learn the opinion word enclosed in more complicated contexts. The performance is improved by using LSTM, attention layer and feature engineering with word distance between surrounding words and target words to produce target-specific memory~\cite{Chen:2017wv}.

\textbf{Aspect-Category Sentiment Analysis}. In this category, the model is asked to predict the sentiment polarity toward a predefined aspect category. Attention-based LSTM with Aspect Embedding~\cite{Wang:2016tf} uses the embedding vectors of aspect words to selectively attend the regions of the representations generated by LSTMs.

\section{Gated Convolutional Network with Aspect Embedding}
In this section, we present a new model for ACSA and ATSA, namely Gated Convolutional network with Aspect Embedding (GCAE), which is more efficient and simpler than recurrent network based models~\cite{Wang:2016tf,Tang:2016th,Ma:2017jo,Chen:2017wv}.
Recurrent neural networks sequentially compose hidden vectors $\mathbf{h}_{i} = f (\mathbf{h}_{i-1})$, which does not enable parallelization over inputs. In the attention layer, softmax normalization also has to wait for all the alignment scores computed by a similarity function. 
Hence, they cannot take advantage of highly-parallelized modern hardware and libraries.
Our model is built on convolutional layers and gating units. Each convolutional filter computes n-gram features at different granularities from the embedding vectors at each position individually. The gating units on top of the convolutional layers at each position are also independent from each other. Therefore, our model is more suitable to parallel computing. 
Moreover, our model is equipped with two kinds of effective filtering mechanisms: the gating units on top of the convolutional layers and the max pooling layer, both of which can accurately generate and select aspect-related sentiment features.

We first briefly review the vanilla CNN for text classification~\cite{Kim:2014vt}. The model achieves state-of-the-art performance on many standard sentiment classification datasets~\cite{Le:2017vg}.

The CNN model consists of an embedding layer, a one-dimension convolutional layer and a max-pooling layer.
The embedding layer takes the indices $w_i \in \{1, 2, \ldots, V\}$ of the input words and outputs the corresponding embedding vectors $\boldsymbol{v}_i \in \mathbb{R}^D$. $D$ denotes the dimension size of the embedding vectors. $V$ is the size of the word vocabulary. The embedding layer is usually initialized with pre-trained embeddings such as GloVe~\cite{Pennington:2014uw}, then they are fine-tuned during the training stage. 
The one-dimension convolutional layer convolves the inputs with multiple convolutional kernels of different widths. Each kernel corresponds a linguistic feature detector which extracts a specific pattern of n-gram at various granularities~\cite{Kalchbrenner:2014wl}.
Specifically, the input sentence is represented by a matrix through the embedding layer, $\mathbf{X} = [\boldsymbol{v}_1, \boldsymbol{v}_2, \ldots, \boldsymbol{v}_L]$, where $L$ is the length of the sentence with padding. A convolutional filter $\mathbf{W}_c \in \mathbb{R}^{D \times k}$ maps $k$ words in the receptive field to a single feature $c$. As we slide the filter across the whole sentence, we obtain a sequence of new features $\mathbf{c} = [c_1, c_2, \ldots, c_L]$.
\begin{equation}
  \label{c4eq:conv}
  c_i = f(\mathbf{X}_{i:i+K} * \mathbf{W}_c + b_c)  \quad,  
\end{equation}
where $b_c \in \mathbb{R}$ is the bias, $f$ is a non-linear activation function such as tanh function, $*$ denotes convolution operation. If there are $n_k$ filters of the same width $k$, the output features form a matrix $\mathbf{C} \in \mathbb{R}^{n_k \times L_k}$. For each convolutional filter, the max-over-time pooling layer takes the maximal value among the generated convolutional features, resulting in a fixed-size vector whose size is equal to the number of filters $n_k$. Finally, a softmax layer uses the vector to predict the sentiment polarity of the input sentence.

\begin{figure}
\centering
  \includegraphics[scale=0.7]{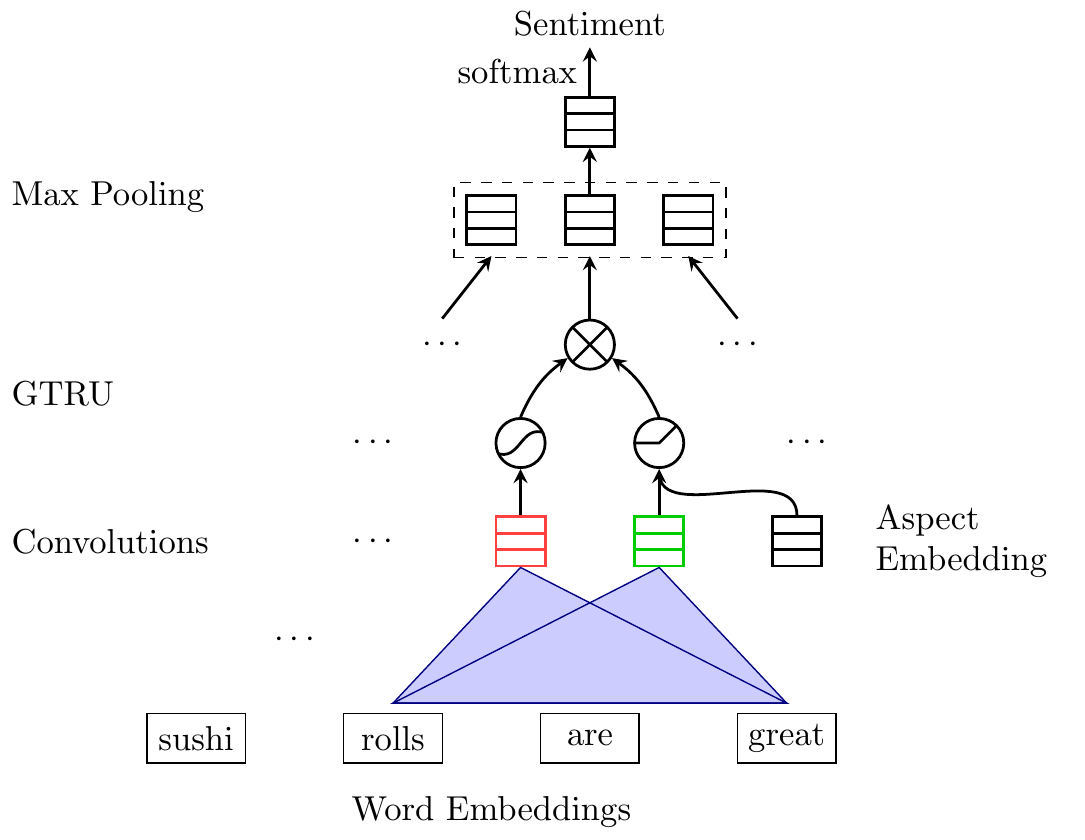}
\caption{Illustration of our model GCAE for ACSA task. A pair of convolutional neuron computes features for a pair of gates: tanh gate and ReLU gate. The ReLU gate receives the given aspect information to control the propagation of sentiment features. The outputs of two gates are element-wisely multiplied for the max pooling layer.}
\label{fig:model}
\end{figure}

Figure~\ref{fig:model} illustrates our model architecture. The Gated Tanh-ReLU Units (GTRU) with aspect embedding are connected to two convolutional neurons at each position $t$. Specifically, we compute the features $c_i$ as
\begin{align}
  \label{eq:gcae_aspect}
  a_i &= \text{relu}(\mathbf{X}_{i:i+k} * \mathbf{W}_a + \mathbf{V}_{a} \boldsymbol{v}_a + b_a)    \\
  \label{eq:gcae_sentiment}
  s_i &= \text{tanh}(\mathbf{X}_{i:i+k} * \mathbf{W}_s + b_s) \\
  \label{eq:gcae_mutiply}
  c_i &= s_i \times a_i \quad ,
\end{align}
where $\boldsymbol{v}_a$ is the embedding vector of the given aspect category in ACSA or computed by another CNN over aspect terms in ATSA. 
The two convolutions in Equation~\ref{eq:gcae_aspect} and \ref{eq:gcae_sentiment} are the same as the convolution in the vanilla CNN, but the convolutional features $a_i$ receives additional aspect information $\boldsymbol{v}_a$ with ReLU activation function. In other words, $s_i$ and $a_i$ are responsible for generating sentiment features and aspect features respectively.  
The above max-over-time pooling layer generates a fixed-size vector $\boldsymbol{e} \in \mathbb{R}^{d_k}$, which keeps the most salient sentiment features of the whole sentence. 
The final fully-connected layer with softmax function uses the vector $\boldsymbol{e}$ to predict the sentiment polarity $\hat{y}$. The model is trained by minimizing the cross-entropy loss between the ground-truth $y$ and the predicted value $\hat{y}$ for all data samples.
\begin{equation}
  \mathcal{L} = - \sum_i \sum_j y_i^j \log \hat{y}_i^j \quad ,
\end{equation}
where $i$ is the index of a data sample, $j$ is the index of a sentiment class.

\section{Gating Mechanisms}
The proposed Gated Tanh-ReLU Units control the path through which the sentiment information flows towards the pooling layer. The gating mechanisms have proven to be effective in LSTM.
In aspect based sentiment analysis, it is very common that different aspects with different sentiments appear in one sentence. The ReLU gate in Equation~\ref{eq:gcae_aspect} does not have upper bound on positive inputs but strictly zero on negative inputs. Therefore, it can output a similarity score according to the relevance between the given aspect information $\boldsymbol{v}_a$ and the aspect feature $a_i$ at position $t$. If this score is zero, the sentiment features $s_i$ would be blocked at the gate; otherwise, its magnitude would be amplified accordingly. The max-over-time pooling further removes the sentiment features which are not significant over the whole sentence.

In language modeling~\cite{Dauphin:2016uja,Kalchbrenner:2016vf,vandenOord:2016tk,Gehring:2017tv}, Gated Tanh Units (GTU) and Gated Linear Units (GLU) have shown effectiveness of gating mechanisms. 
GTU is represented by $\tanh(\mathbf{X}*\mathbf{W} + b) \times \sigma (\mathbf{X}*\mathbf{V} + c)$, in which the sigmoid gates control features for predicting the next word in a stacked convolutional block. 
To overcome the gradient vanishing problem of GTU, GLU uses $(\mathbf{X}*\mathbf{W} + b) \times \sigma (\mathbf{X}*\mathbf{V} + c)$ instead, so that the gradients would not be downscaled to propagate through many stacked convolutional layers. 
However, a neural network that has only one convolutional layer would not suffer from gradient vanish problem during training.
We show that on text classification problem, our GTRU is more effective than these two gating units. 

\section{GCAE on ATSA}
\begin{figure*}
\centering
  \includegraphics[scale=0.85]{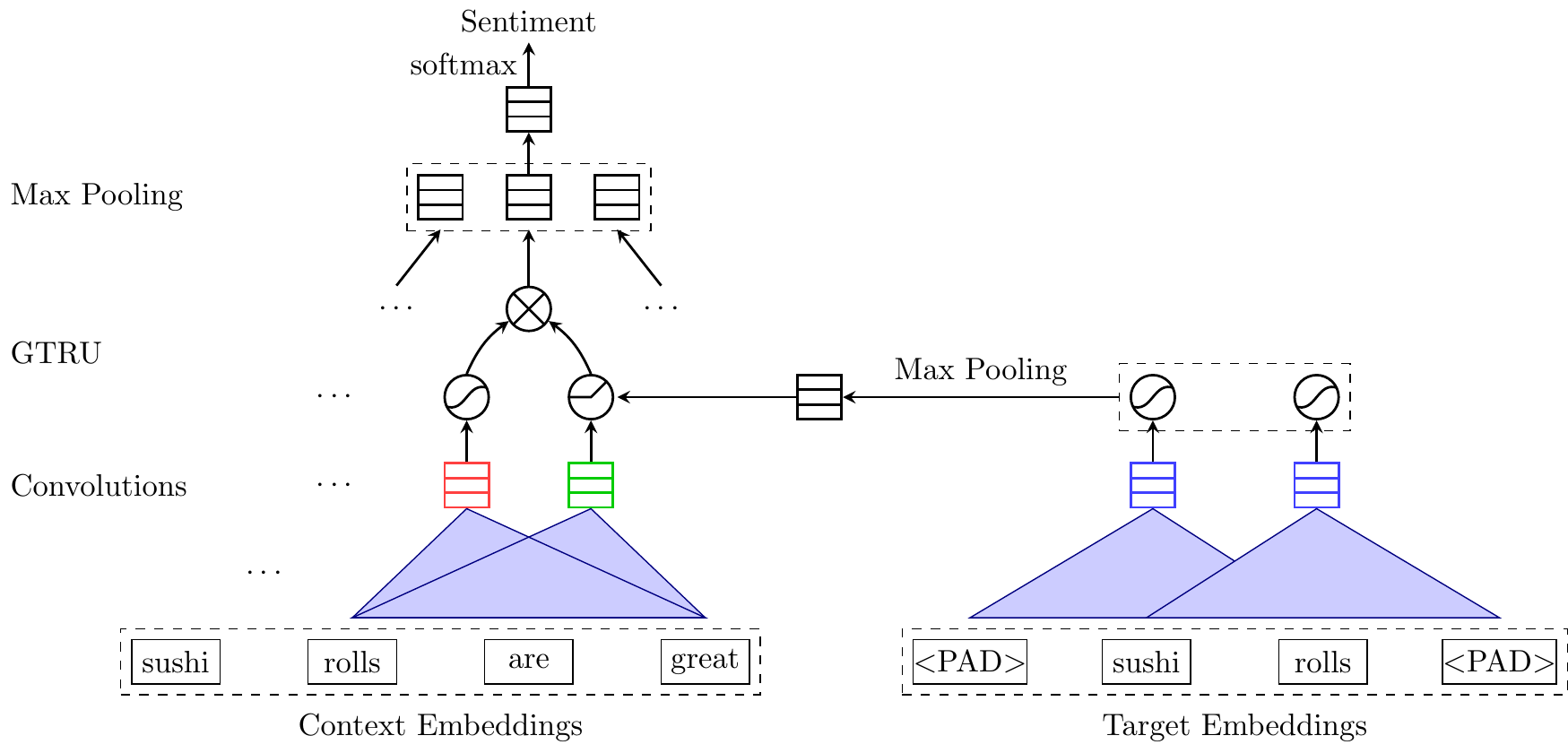}
\caption{Illustration of model GCAE for ATSA task. It has an additional convolutional layer on aspect terms.}
\label{fig:gcae_atsa}
\end{figure*}

ATSA task is defined to predict the sentiment polarity of the aspect terms in the given sentence.
We simply extend GCAE by adding a small convolutional layer on aspect terms, as shown in Figure~\ref{fig:gcae_atsa}. In ACSA, the aspect information controlling the flow of sentiment features in GTRU is from one aspect word; while in ATSA, such information  is provided by a small CNN on aspect terms $[w_i, w_{i+1}, \ldots, w_{i+k}]$. The additional CNN extracts the important features from multiple words while retains the ability of parallel computing. 

\section{Experiments}

\subsection{Datasets and Experiment Preparation}
We conduct experiments on public datasets from SemEval workshops~\cite{Pontiki:2014ex}, which consist of customer reviews about restaurants and laptops. 
Some existing work~\cite{Wang:2016tf,Ma:2017jo,Chen:2017wv} removed ``conflict'' labels from four sentiment labels, which makes their results incomparable to those from the workshop report~\cite{Kiritchenko:2014jw}.  We reimplemented the compared methods, and used hyper-parameter settings described in these references.

The sentences which have different sentiment labels for different aspects or targets in the sentence are more common in review data than in standard sentiment classification benchmark. The sentence in Table~\ref{tbl:testM} shows the reviewer's different attitude towards two aspects: food and delivery. Therefore, to access how the models perform on review sentences more accurately, we create small but difficult datasets, which are made up of the sentences having opposite or different sentiments on different aspects/targets. In Table~\ref{tbl:testM}, the two identical sentences but with different sentiment labels are both included in the dataset. If a sentence has 4 aspect targets, this sentence would have 4 copies in the data set, each of which is associated with different target and sentiment label.  

\begin{table*}[h]
\centering
\begin{tabular}{l|c|c}
\hline
Sentence & aspect category/term & sentiment label \\
\hline
Average to good Thai food, but terrible delivery. & food & positive \\
Average to good Thai food, but terrible delivery. & delivery & negative \\
\hline
\end{tabular}
\caption{Two example sentences in one hard test set of restaurant review dataset of SemEval 2014.}
\label{tbl:testM}
\end{table*}

For ACSA task, we conduct experiments on restaurant review data of SemEval 2014 Task 4. There are 5 aspects: food, price, service, ambience, and misc; 4 sentiment polarities: positive, negative, neutral, and conflict. 
By merging restaurant reviews of three years 2014 - 2016, we obtain a larger dataset called ``Restaurant-Large''. Incompatibilities of data are fixed during merging. We replace conflict labels with neutral labels in the 2014 dataset. In the 2015 and 2016 datasets, there could be multiple pairs of ``aspect terms'' and ``aspect category'' in one sentence. For each sentence, let $p$ denote the number of positive labels minus the number of negative labels. We assign a sentence a positive label if $p > 0$, a negative label if $p<0$, or a neutral label if $p=0$. After removing duplicates, the statistics are show in Table~\ref{tbl:r14r16}. The resulting dataset has 8 aspects: {restaurant}, {food}, {drinks}, {ambience}, {service}, {price}, {misc} and {location}.

For ATSA task, we use restaurant reviews and laptop reviews from SemEval 2014 Task 4.
On each dataset, we duplicate each sentence $n_a$ times, which is equal to the number of associated aspect categories (ACSA) or aspect terms (ATSA)~\cite{Ruder:2016ve,Ruder:2016ug}. The statistics of the datasets are shown in Table~\ref{tbl:r14r16}.

The sizes of hard data sets are also shown in Table~\ref{tbl:r14r16}. The test set is designed to measure whether a model can detect multiple different sentiment polarities in one sentence toward different entities. Without such sentences, a classifier for overall sentiment classification might be good enough for the sentences associated with only one sentiment label.  

\begin{table*}[ht]
\centering
\begin{tabular}{l|ll|ll|ll|ll}
\hline
\multirow{2}{*}{} & \multicolumn{2}{c|}{Positive} & \multicolumn{2}{c|}{Negative} & \multicolumn{2}{c|}{Neutral} & \multicolumn{2}{c}{Conflict} \\ 
\cline{2-9} 
                 & Train      & Test      & Train      & Test      & Train      & Test      & Train      & Test \\ \hline
Restaurant-Large & 2710       & 1505      & 1198       & 680       & 757        & 241       & -          & -    \\
Restaurant-Large-Hard & 182   & 92        & 178        & 81        & 107        & 61        & -          & -    \\ \hline
Restaurant-2014  & 2179    & 657   & 839   & 222  & 500   & 94   & 195    & 52   \\ 
Restaurant-2014-Hard  & 139   & 32        & 136        & 26        & 50         & 12        & 40         & 19   \\
\hline
\end{tabular}
\caption{Statistics of the datasets for ACSA task. The hard dataset is only made up of sentences having multiple aspect labels associated with multiple sentiments.}
\label{tbl:r14r16}
\end{table*}

\begin{table*}[ht]
\centering
\begin{tabular}{l|ll|ll|ll|ll}
\hline
\multirow{2}{*}{} & \multicolumn{2}{c|}{Positive} & \multicolumn{2}{c|}{Negative} & \multicolumn{2}{c|}{Neutral} & \multicolumn{2}{c}{Conflict}  \\ \cline{2-9} 
                 & Train   & Test  & Train & Test & Train & Test & Train  & Test \\ \hline
Restaurant       & 2164    & 728   & 805   & 196  & 633   & 196   & 91    & 14   \\ 
Restaurant-Hard  & 379     & 92    & 323   & 62   & 293   & 83   & 43     & 8    \\ \hline
Laptop           & 987     & 341   & 866   & 128  & 460   & 169  & 45     & 16   \\
Laptop-Hard      & 159     & 31    & 147   & 25   & 173   & 49   & 17     & 3    \\
\hline
\end{tabular}
\caption{Statistics of the datasets for ATSA task.}
\label{tbl:atsa2014}
\end{table*}

In our experiments, word embedding vectors are initialized with 300-dimension GloVe vectors which are pre-trained on unlabeled data of 840 billion tokens~\cite{Pennington:2014uw}. Words out of the vocabulary of GloVe are randomly initialized with a uniform distribution $U(-0.25, 0.25)$.
We use Adagrad~\cite{Duchi:2011wu} with a batch size of 32 instances, default learning rate of $1e-2$, and maximal epochs of 30. We only fine tune early stopping with 5-fold cross validation on training datasets. All neural models are implemented in PyTorch.

\subsection{Compared Methods}
To comprehensively evaluate the performance of GCAE, we compare our model against the following models. 

\textbf{NRC-Canada}~\cite{Kiritchenko:2014jw} is the top method in SemEval 2014 Task 4 for ACSA and ATSA task. SVM is trained with extensive feature engineering: various types of n-grams, POS tags, and lexicon features. The sentiment lexicons improve the performance significantly, but it requires large scale labeled data: 183 thousand Yelp reviews, 124 thousand Amazon laptop reviews, 56 million tweets, and 3 sentiment lexicons labeled manually.

\textbf{CNN}~\cite{Kim:2014vt} is widely used on text classification task. It cannot directly capture aspect-specific sentiment information on ACSA task, but it provides a very strong baseline for sentiment classification. We set the widths of filters to 3, 4, 5 with 100 features each.

\textbf{TD-LSTM}~\cite{Tang:2016th} uses two LSTM networks to model the preceding and following contexts of the target to generate target-dependent representation for sentiment prediction.

\textbf{ATAE-LSTM}~\cite{Wang:2016tf} is an attention-based LSTM for ACSA task. It appends the given aspect embedding with each word embedding as the input of LSTM, and has an attention layer above the LSTM layer. 

\textbf{IAN}~\cite{Ma:2017jo} stands for interactive attention network for ATSA task, which is also based on LSTM and attention mechanisms. 

\textbf{RAM}~\cite{Chen:2017wv} is a recurrent attention network for ATSA task, which uses LSTM and multiple attention mechanisms.

\textbf{GCN} stands for gated convolutional neural network, in which GTRU does not have the aspect embedding as an additional input.

\subsection{Results and Analysis}
\subsubsection{ACSA}
\begin{table*}[ht]
\centering
\begin{tabular}{l|ll|ll}
\hline
\multicolumn{1}{c|}{\multirow{2}{*}{Models}} & \multicolumn{2}{c|}{Restaurant-Large}               & \multicolumn{2}{c}{Restaurant 2014}                        \\ \cline{2-5} 
\multicolumn{1}{c|}{}                    & \multicolumn{1}{c}{Test} & \multicolumn{1}{c|}{Hard Test} & \multicolumn{1}{c}{Test} & \multicolumn{1}{c}{Hard Test}        \\ \hline
SVM*             & -                     & -                     & 75.32                 & -          \\
SVM + lexicons*  & -                   & -                     & \textbf{82.93}        & -          \\ \hline
ATAE-LSTM        & 83.91$\pm$0.49      & 66.32$\pm$2.28        & 78.29$\pm$0.68     & 45.62$\pm$0.90                 \\
CNN              & 84.28$\pm$0.15      & 50.43$\pm$0.38        & 79.47$\pm$0.32     & 44.94$\pm$0.01                 \\
GCN              & 84.48$\pm$0.06      & 50.08$\pm$0.31        & \textbf{79.67$\pm$0.35}   & 44.49$\pm$1.52          \\
GCAE             & \textbf{85.92$\pm$0.27}     & \textbf{70.75$\pm$1.19}      & 79.35$\pm$0.34    & \textbf{50.55$\pm$1.83} \\ \hline
\end{tabular}
\caption{The accuracy of all models on test sets and on the subsets made up of test sentences that have multiple sentiments and multiple aspect terms. Restaurant-Large dataset is created by merging all the restaurant reviews of SemEval workshops within three years. `*': the results with SVM are retrieved from NRC-Canada~\cite{Kiritchenko:2014jw}. }
\label{tbl:rest_lap}
\end{table*}

Following the SemEval workshop, we report the overall accuracy of all competing models over the test datasets of restaurant reviews  as well as the hard test datasets. Every experiment is repeated five times. The mean and the standard deviation are reported in Table~\ref{tbl:rest_lap}.

LSTM based model ATAE-LSTM has the worst performance of all neural networks. Aspect-based sentiment analysis is to extract the sentiment information closely related to the given aspect. It is important to separate aspect information and sentiment information from the extracted information of sentences. The context vectors generated by LSTM have to convey the two kinds of information at the same time. Moreover, the attention scores generated by the similarity scoring function are for the entire context vector. 

GCAE improves the performance by 1.1\% to 2.5\% compared with ATAE-LSTM. 
First, our model incorporates GTRU to control the sentiment information flow according to the given aspect information at each dimension of the context vectors. The element-wise gating mechanism works at fine granularity instead of exerting an alignment score to all the dimensions of the context vectors in the attention layer of other models.
Second, GCAE does not generate a single context vector, but two vectors for aspect and sentiment features respectively, so that aspect and sentiment information is unraveled. 
By comparing the performance on the hard test datasets against CNN, it is easy to see the convolutional layer of GCAE is able to differentiate the sentiments of multiple entities. 

Convolutional neural networks CNN and GCN are not designed for aspect based sentiment analysis, but their performance exceeds that of ATAE-LSTM.

The performance of SVM~\cite{Kiritchenko:2014jw} depends on the availability of the features it can use. Without 
the large amount of sentiment lexicons, SVM perform worse than neural methods. With multiple sentiment lexicons, the performance is increased by 7.6\%.  This inspires us to work on leveraging sentiment lexicons in neural networks in the future.

The hard test datasets consist of replicated sentences with different sentiments towards different aspects. 
The models which cannot utilize the given aspect information such as CNN and GCN perform poorly as expected, but GCAE  has higher accuracy than other neural network models. GCAE achieves  4\% higher accuracy than ATAE-LSTM on Restaurant-Large and 5\% higher on SemEval-2014 on ACSA task. However, GCN, which does not have aspect modeling part, has higher score than GCAE on the original restaurant dataset. It suggests that GCN performs better than GCAE when there is only one sentiment label in the given sentence, but not on the hard test dataset. 

\subsubsection{ATSA}
\begin{table*}[ht]
\centering
\begin{tabular}{l|ll|ll}
\hline
\multirow{2}{*}{Models} & \multicolumn{2}{c|}{Restaurant}   & \multicolumn{2}{c}{Laptop}             \\ \cline{2-5} 
& \multicolumn{1}{c}{Test} & \multicolumn{1}{c|}{Hard Test} & \multicolumn{1}{c}{Test} & \multicolumn{1}{c}{Hard Test}     \\ \hline
SVM*                  &  77.13            & -          &    63.61          & -                \\
SVM + lexicons*         &  \textbf{80.16}   & -          &  \textbf{70.49}   & -                \\ \hline
TD-LSTM               & 73.44$\pm$1.17    & 56.48$\pm$2.46    & 62.23$\pm$0.92    & 46.11$\pm$1.89   \\
ATAE-LSTM             & 73.74$\pm$3.01    & 50.98$\pm$2.27    & 64.38$\pm$4.52    & 40.39$\pm$1.30   \\
IAN                   & 76.34$\pm$0.27    & 55.16$\pm$1.97    & 68.49$\pm$0.57    & 44.51$\pm$0.48   \\
RAM                   & 76.97$\pm$0.64    & 55.85$\pm$1.60    & 68.48$\pm$0.85    & 45.37$\pm$2.03   \\
GCAE                  & \textbf{77.28$\pm$0.32}        & \textbf{56.73$\pm$0.56}         & \textbf{69.14$\pm$0.32}                      & \textbf{47.06$\pm$2.45}        \\ \hline
\end{tabular}
\caption{The accuracy of ATSA subtask on SemEval 2014 Task 4. `*': the results with SVM are retrieved from NRC-Canada~\cite{Kiritchenko:2014jw}}
\label{tbl:atsa}
\end{table*}
We apply the extended version of GCAE on ATSA task. On this task, the aspect terms are marked in the sentences and usually consist of multiple words.
We compare IAN~\cite{Ma:2017jo}, RAM~\cite{Chen:2017wv}, TD-LSTM~\cite{Tang:2016th}, ATAE-LSTM~\cite{Wang:2016tf}, and our GCAE model in Table~\ref{tbl:atsa}. The models other than GCAE is based on LSTM and attention mechanisms. 
IAN has better performance than TD-LSTM and ATAE-LSTM, because two attention layers guides the representation learning of the context and the entity interactively. RAM also achieves good accuracy by combining multiple attentions with a recurrent neural network, but it needs more training time as shown in the following section.
On the hard test dataset, GCAE has 1\% higher accuracy than RAM on restaurant data and 1.7\% higher on laptop data. 
GCAE uses the outputs of the small CNN over aspect terms to guide the composition of the sentiment features through the ReLU gate. Because of the gating mechanisms and the convolutional layer over aspect terms, GCAE outperforms other neural models and basic SVM. 
Again, large scale sentiment lexicons bring significant improvement to SVM.



\subsection{Training Time}
\begin{table}
\centering
\begin{tabular}{l|r}
\hline
Model      & ATSA   \\ \hline
ATAE       & 25.28  \\
IAN        & 82.87  \\
RAM        & 64.16  \\
TD-LSTM    & 19.39  \\
GCAE      & 3.33   \\ \hline
\end{tabular}
\caption{The time to converge in seconds on ATSA task.}
\label{tbl:acsa_time}
\end{table}

We record the training time of all models until convergence on a validation set on a desktop machine with a 1080 Ti GPU, as shown in Table~\ref{tbl:acsa_time}.
LSTM based models take more training time than convolutional models. On ATSA task, because of multiple attention layers in IAN and RAM, they need even more time to finish the training.
GCAE is much faster than other neural models, because neither convolutional operation nor GTRU has time dependency compared with LSTM and attention layer. Therefore, it is easier for hardware and libraries to parallel the computing process.
Since the performance of SVM is retrieved from the original paper, we are not able to compare the training time of SVM.

\subsection{Gating Mechanisms}
\begin{table}[]
\centering
\begin{tabular}{l|ll|ll}
\hline
\multirow{2}{*}{Gates}  & \multicolumn{2}{c|}{Restaurant-Large}   & \multicolumn{2}{c}{Restaurant 2014} \\ \cline{2-5} 
& \multicolumn{1}{c}{Test} & \multicolumn{1}{c|}{Hard Test} & \multicolumn{1}{c}{Test} & \multicolumn{1}{c}{Hard Test}  \\ \hline
GTU                   &  84.62    &  60.25    & 79.31    & \textbf{51.93}       \\
GLU                   &  84.74    &  59.82    & 79.12    & 50.80                \\
GTRU                  &  \textbf{85.92}    &  \textbf{70.75}    & \textbf{79.35}        & 50.55   \\ \hline
\end{tabular}
\caption{The accuracy of different gating units on restaurant reviews on ACSA task.}
\label{tbl:gates}
\end{table}

In this section, we compare 
GLU $(\mathbf{X}*\mathbf{W} + b) \times \sigma (\mathbf{X}*\mathbf{W}_a + \mathbf{V}\boldsymbol{v}_a + b_a)$ ~\cite{Dauphin:2016uja}, 
GTU $\tanh(\mathbf{X}*\mathbf{W} + b) \times \sigma (\mathbf{X}*\mathbf{W}_a + \mathbf{V}\boldsymbol{v}_a + b_a)$ ~\cite{vandenOord:2016tk}, 
and GTRU used in GCAE. 
Table~\ref{tbl:gates} shows that all of three gating units achieve relatively high accuracy on restaurant datasets. GTRU outperforms the other gates. It has a convolutional layer generating aspect features via ReLU activation function, which controls the magnitude of the sentiment signals according to the given aspect information. On the other hand, the sigmoid function in GTU and GLU has the upper bound $+1$, which may not be able to distill sentiment features effectively.

\section{Visualization}
\begin{figure}
  \centering
  \includegraphics[scale=0.8]{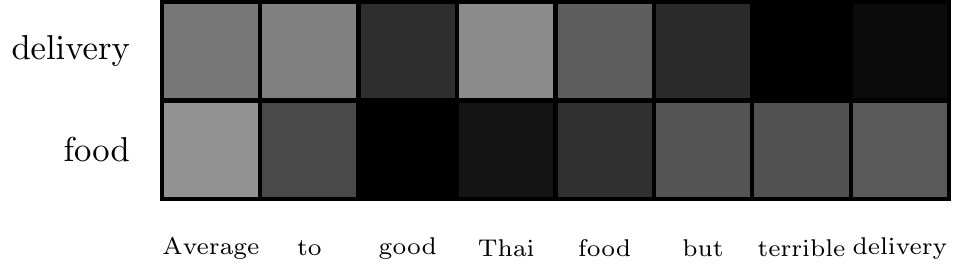}
  \caption{The outputs of the ReLU gates in GTRU.} 
  \label{fig:vis}
\end{figure}
In this section, we take a concrete review sentence as an example to illustrate how the proposed gate GTRU works. It is more difficult to visualize the weights generated by the gates than the attention weights in other neural networks. The attention weight score is a global score over the words and the vector dimensions; whereas in our model, there are $N_{\text{word}} \times N_{\text{filter}} \times N_{\text{dimension}}$ gate outputs. Therefore, we train a small model with only one filter which is only three word wide. Then, for each word, we sum the $N_{\text{dimension}}$ outputs of the ReLU gates. After normalization, we plot the values on each word in Figure~\ref{fig:vis}. Given different aspect targets, the ReLU gates would control the magnitude of the outputs of the tanh gates.

\section{Conclusions and Future Work}
In this paper, we proposed an efficient convolutional neural network with gating mechanisms for ACSA and ATSA tasks. 
GTRU can effectively control the sentiment flow according to the given aspect information, and two convolutional layers model the aspect and sentiment information separately.
We prove the performance improvement compared with other neural models by extensive experiments on SemEval datasets. 
How to leverage large-scale sentiment lexicons in neural networks would be our future work.

\bibliography{acl2018}
\bibliographystyle{acl_natbib}

\end{document}